\documentclass[runningheads]{llncs}
\usepackage[T1]{fontenc}
\usepackage{graphicx}
\usepackage{amsmath,amssymb}
\usepackage{algorithm}
\usepackage{algpseudocode}
\usepackage{float}
\usepackage{placeins}
\usepackage{pifont}
\usepackage[table]{xcolor}
\usepackage{array}
\usepackage{tabularx}
\definecolor{TableHeader}{HTML}{EAF1F6}
\definecolor{TableStripe}{HTML}{F7F9FB}
\definecolor{AblationBg}{HTML}{FBF4E7}
\definecolor{OursBg}{HTML}{FBEDEE}
\newcolumntype{Y}{>{\centering\arraybackslash}X}
\begin{document}
\raggedbottom
\title{Decision-Oriented Uncertainty Quantification for Risk Control in Earth System Spatiotemporal Foundation Models}
\titlerunning{Decision-Oriented UQ for Earth System ST-FMs}
%
\author{Ji Lu\inst{1} \and
Huiran Duan\inst{2} \and
Bo Zhao\inst{3} \and
Xianglong Wang\inst{4} \and
Yiru Fang\inst{4} \and
Kuo Yang\inst{5} \and
Xiaoqin Feng\inst{6} \and
Jianping Gou\inst{7}\thanks{Corresponding author.}}
\authorrunning{J. Lu et al.}
%
\institute{Vanderbilt University, United States\\
\and
City University of New York, United States\\
\and
Yale University, United States\\
\and
Wyze Inc., United States\\
\and
Northeastern University, United States\\
\and
University of Southern California, United States\\
\and
Southwest University, China\quad
\email{pjgzy61@swu.edu.cn}}
\maketitle              
\begin{abstract}
Earth system modeling is shifting from task-specific predictors toward
foundation models with general spatiotemporal representation capabilities.
Although these models can jointly encode dynamic Earth fields, external
forcings, and static geographic context for multistep forecasting, accurate
point predictions or statistically calibrated intervals alone are insufficient
for high-impact applications such as extreme-weather warning, flood control,
renewable-energy dispatch, and emergency resource allocation. What matters in
practice is whether predictive uncertainty can be translated into reliable
decision risk under specific actions, loss functions, and risk preferences.
We propose a decision-oriented uncertainty quantification framework for Earth
system spatiotemporal foundation models. The framework produces predictive
distributions of future states and uses a decision risk adapter to map forecast
samples, decision context, and utility functions into action-conditional risks.
A utility-aware calibration module further enforces reliability at the
downstream decision-loss level rather than only at the forecast-value level.
Calibrated risks are then used to select warning, dispatch, inspection, or
resource-allocation actions. Compared with the strongest baseline, the proposed method reduces decision
regret by \textbf{18.7\%}, lowers the missed-event rate from \textbf{14.2\%}
to \textbf{9.1\%}, and improves expected utility by \textbf{11.6\%}, while
maintaining \textbf{90.4\%} predictive coverage and reducing decision
calibration error from \textbf{0.083} to \textbf{0.047}. These results suggest
that decision-oriented uncertainty quantification can improve the robustness
and operational value of Earth system foundation models in risk-sensitive
applications.

\keywords{Earth system modeling \and Spatiotemporal foundation model \and
Uncertainty quantification \and Decision risk \and Utility-aware calibration}
\end{abstract}
\section{Introduction}

Recently, deep learning has applied in many fields~\cite{li2025frequency,11541222,li2026comprehensive,li2025ddtime}. Advances in multimodal and high-performance AI have been particularly rapid.~\cite{Li2025Efficient,zhao2026mis,li2025pruning,li2023less,li2024neural,gou2021knowledge,feng2026mpq,feng2026s} Many scientific computing approaches can be further advanced by integrating deep learning~\cite{cheng2025cgmatch,qi2026next,gou2022multilevel,gou2025multi}. Earth system forecasting provides essential scientific support for
extreme-weather warnings, flood and drought management, renewable-energy
operations, and emergency response. Recent data-driven systems such as
Pangu-Weather and GraphCast have demonstrated that learned global models can
deliver accurate medium-range forecasts at substantially lower inference cost
than conventional numerical pipelines~\cite{bi2023pangu,lam2023graphcast,li2026rethinking,wu2026roboalign,11460474,xie2026symmetry}.
The field is now moving beyond task-specific predictors toward foundation
models with transferable spatiotemporal representations. Aurora, for example,
is pretrained on heterogeneous geophysical data and adapted to weather, air
quality, ocean-wave, and tropical-cyclone forecasting~\cite{bodnar2025aurora}.
More recently, the Earth System Foundation Model (ESFM) has extended this
direction by integrating dense reanalysis, sparse station observations, and
satellite data within a unified architecture~\cite{ozdemir2026esfm}. These
developments build on broader progress in versatile physical-dynamics modeling
with EarthFarseer~\cite{wu2024earthfarsser}, unified global--regional weather
prediction with OneForecast~\cite{gao2025oneforecast}, stable long-horizon
forecasting with TritonCast~\cite{wu2025tritoncast}, and error correction for
coupled dynamical systems with PnP-Corrector~\cite{wu2026pnp}. Together, they
create a promising basis for multivariable, multiregional, and multihorizon
Earth system prediction.

The scope of data-driven Earth system modeling has also expanded beyond the
atmosphere. AI-GOMS introduced a transferable backbone and downstream-task
paradigm for
global ocean prediction~\cite{xiong2023aigoms}, while NeuralOM targets stable
S2S ocean simulation with physics-guided graph
operators~\cite{gao2026neuralom}. Physics-guided models have further improved
the prediction of global extreme marine heatwaves~\cite{shu2025mhw} and
extended-range precipitation over East Asia~\cite{wang2026precip}. These
advances broaden the range of high-impact applications served by Earth system
models, while also increasing the importance of representing uncertainty
consistently across variables, regions, and forecast horizons.

Higher predictive accuracy, however, does not necessarily yield better
operational decisions. In extreme-precipitation warning, a missed event can
cause severe human and economic losses, whereas repeated false alarms consume
resources and erode public trust. In reservoir operation, an early release may
reduce flood risk but increase the probability of subsequent water shortage.
Similar asymmetries arise in renewable-energy dispatch and emergency resource
allocation. Moreover, the same predictive distribution may imply different
risks under different actions, cost structures, and risk preferences.
Operational systems must therefore answer not only \emph{what is likely to
happen}, but also \emph{what loss each action may incur under an uncertain
future}.

Uncertainty quantification (UQ) offers tools for representing such uncertain
futures. Probabilistic weather models such as GenCast generate ensembles of
plausible trajectories rather than a single deterministic forecast
~\cite{price2025gencast}, while Tyche uses one-step conditional flow matching
for efficient probabilistic forecasting~\cite{xu2026tyche}. Recent work has further explored lightweight
stochastic attention for calibrating scientific foundation models
~\cite{yadav2026stochastic}, empirical neural tangent kernels for scalable
extreme-weather UQ~\cite{minoza2026ntkuq}, and online conformal prediction for
coverage guarantees in probabilistic AI weather forecasts
~\cite{asch2026conformal}. Conformal methods provide a particularly general
route to distribution-free predictive coverage~\cite{angelopoulos2023conformal}.
Nevertheless, these approaches are usually assessed through forecast-level
criteria such as likelihood, continuous ranked probability score, interval
coverage, sharpness, or expected calibration error.

Forecast-level statistical reliability is not equivalent to decision-level
risk reliability. A prediction interval with correct marginal coverage can
still induce costly actions under tail events, distribution shift, or regional
heterogeneity. This mismatch becomes especially consequential when the costs
of misses and false alarms are strongly asymmetric. Recent decision-aligned UQ
research has likewise shown that generic uncertainty metrics need not rank
models according to their realized downstream utility
~\cite{schneider2026decision}. These observations expose a central gap:
existing methods primarily calibrate predicted values or probabilities, while
operational users require reliable estimates of the losses induced by
candidate actions. Spatiotemporal Forecasting as Planning takes an important
step toward task-aware prediction by using a generative world model and
domain-specific rewards~\cite{wu2025planning}; our focus is complementary,
targeting the reliability of action-conditional risks produced from predictive
uncertainty.

To bridge this gap, we propose a decision-oriented UQ framework for Earth
system spatiotemporal foundation models, as illustrated in Fig.~\ref{fig1}.
The foundation-model backbone jointly encodes historical Earth system states,
future external forcings, and static geographic information to produce a
predictive distribution over future spatiotemporal states. A decision risk
adapter then combines forecast samples with candidate actions, decision
context, and utility functions to estimate action-conditional risks. Finally,
a utility-aware calibration module targets reliability at the downstream loss
level rather than only at the forecast-value level. The calibrated risks
support warning, dispatch, inspection, and resource-allocation decisions,
thereby connecting probabilistic forecasting directly to risk-aware action
selection.

We evaluate the proposed framework in multiple high-impact Earth system
decision settings using both predictive and decision-centric criteria,
including predictive coverage, decision calibration error, decision regret,
missed- and false-event rates, selective risk, and expected utility.
Compared with the strongest baseline, our method reduces decision regret by
18.7\%, lowers the missed-event rate from 14.2\% to 9.1\%, and improves
expected utility by 11.6\%. Meanwhile, it maintains 90.4\% predictive coverage
and reduces decision calibration error from 0.083 to 0.047. These results
indicate that explicitly modeling and calibrating action-conditional risk can
improve decision quality without sacrificing forecast reliability.

Our main contributions are threefold:
\begin{enumerate}
\item We formulate decision-oriented UQ for Earth system spatiotemporal
foundation models, extending the objective from forecast-level statistical
reliability to action-conditional decision-risk reliability.
\item We introduce a unified framework comprising predictive distributions, a
decision risk adapter, and utility-aware calibration, enabling an explicit
transformation from forecast uncertainty to action-conditional risk under
different loss functions and risk preferences.
\item We establish an evaluation protocol that combines forecast quality with
decision regret, missed-event rate, selective risk, and expected utility,
providing a decision-centric assessment for high-impact Earth system tasks.
\end{enumerate}

\section{Related Work}

\subsection{Earth System Spatiotemporal Forecasting}

Pangu-Weather and GraphCast advance data-driven global weather forecasting
~\cite{bi2023pangu,lam2023graphcast,li2026mrmad,li2026diff}. EarthFarseer strengthens local and global
spatiotemporal modeling~\cite{wu2024earthfarsser}, while OneForecast unifies
global and regional prediction~\cite{gao2025oneforecast}. Aurora and ESFM use
heterogeneous Earth data to support weather, ocean, air-quality, and
extreme-event forecasting~\cite{bodnar2025aurora,ozdemir2026esfm}.

TritonCast and PnP-Corrector improve long-term stability through small-scale
process modeling and error correction~\cite{wu2025tritoncast,wu2026pnp}.
AI-GOMS and NeuralOM extend data-driven modeling to global ocean prediction
~\cite{xiong2023aigoms,gao2026neuralom}. These methods primarily optimize
forecast accuracy and simulation stability, but do not explicitly represent
the decision risk induced by predictive uncertainty.

\subsection{Uncertainty Quantification for Earth System Forecasting}

GenCast generates ensembles of plausible weather trajectories
~\cite{price2025gencast}, and Tyche uses a one-step conditional flow to improve
sampling efficiency~\cite{xu2026tyche}. Stochastic attention and empirical
neural tangent kernels provide lightweight uncertainty estimates for
scientific foundation models and extreme-weather forecasting
~\cite{yadav2026stochastic,minoza2026ntkuq}.

Conformal prediction provides coverage guarantees under weak distributional
assumptions and applies to probabilistic AI weather models
~\cite{angelopoulos2023conformal,asch2026conformal}. Existing methods mainly
evaluate UQ with CRPS, coverage, interval width, and calibration error. Such
forecast-level calibration does not directly guarantee reliable decision risk
under different actions, costs, and risk preferences.

\subsection{Decision-Oriented Uncertainty Quantification}

Decision-focused learning optimizes downstream task loss rather than prediction
error alone~\cite{wilder2019dfl}. Spatiotemporal Forecasting as Planning uses a
generative world model and task rewards to search for high-value future states
~\cite{wu2025planning}. Decision-aligned UQ further shows that generic metrics
such as NLL and ECE do not necessarily reflect realized downstream utility
~\cite{schneider2026decision}.

Conformal risk control and selective prediction provide foundations for
controlling downstream losses~\cite{angelopoulos2024crc}. However, most studies
focus on general classification or regression and do not address high-dimensional
spatiotemporal outputs, multistep error propagation, and extreme events. Our
framework maps forecast samples, actions, and utility functions to
action-conditional risks and calibrates reliability directly at the
decision-loss level.

\section{Method}

\begin{figure}[H]
\centering
\includegraphics[width=\textwidth]{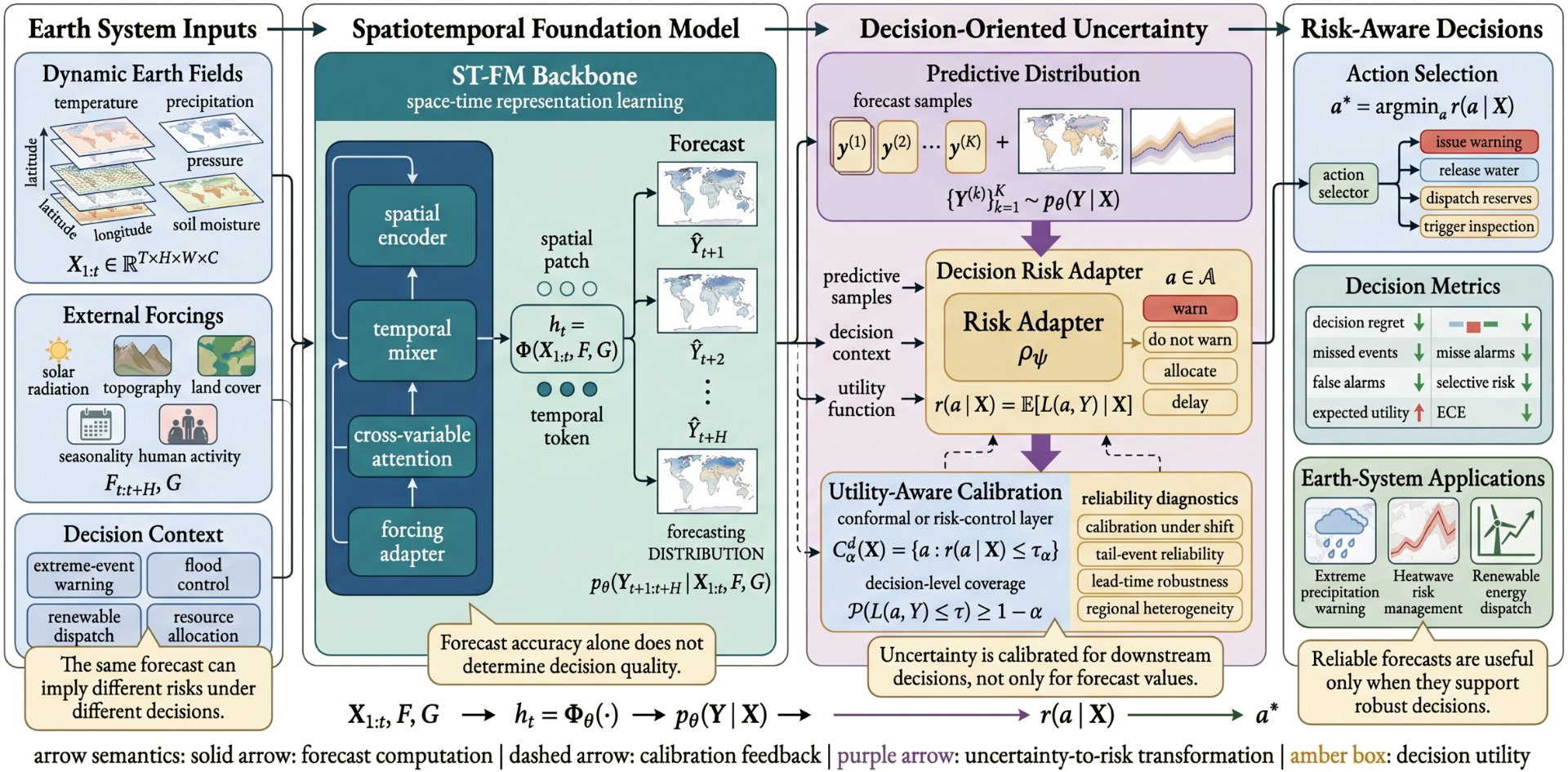}
\caption{Overview of the proposed decision-oriented uncertainty quantification
framework. The spatiotemporal foundation model produces forecast samples, the
risk adapter maps them to action-conditional risks, and utility-aware
calibration supports risk-aware action selection.}
\label{fig1}
\end{figure}

\subsection{Problem Formulation}

Let $\mathbf{X}_{1:t}$ denote historical multivariate Earth fields,
$\mathbf{F}_{t+1:t+H}$ future external forcings, $\mathbf{G}$ static geographic
attributes, and $\mathbf{C}$ the decision context. We write
$\mathbf{I}=(\mathbf{X}_{1:t},\mathbf{F}_{t+1:t+H},\mathbf{G})$ for the
forecast information and $\mathbf{Y}=\mathbf{Y}_{t+1:t+H}$ for the future
Earth state. Given a finite action set $\mathcal{A}$ and a context-dependent
loss $L(a,\mathbf{Y};\mathbf{C})$, the optimal decision minimizes conditional
risk:
\begin{equation}
a^\star
=\arg\min_{a\in\mathcal{A}} r(a\mid\mathbf{I},\mathbf{C}),\qquad
r(a\mid\mathbf{I},\mathbf{C})
=\mathbb{E}\!\left[L(a,\mathbf{Y};\mathbf{C})\mid
\mathbf{I},\mathbf{C}\right].
\label{eq:decision_objective}
\end{equation}
The loss captures both action cost and outcome severity; utility maximization
is equivalent under $U=-L$. Our goal is to estimate and calibrate
$r(a\mid\mathbf{I},\mathbf{C})$, rather than only calibrating individual
forecast variables.

\subsection{Probabilistic Spatiotemporal Foundation Model}

As shown in Fig.~\ref{fig1}, the backbone combines a spatial encoder, a
temporal mixer, cross-variable attention, and a forcing adapter. It produces
the latent state and predictive distribution
\begin{equation}
\mathbf{h}_t=\Phi_\theta(\mathbf{X}_{1:t},
\mathbf{F}_{t+1:t+H},\mathbf{G}),\qquad
p_\theta(\mathbf{Y}\mid\mathbf{I})=
D_\theta(\mathbf{h}_t).
\label{eq:forecast_distribution}
\end{equation}
We train the backbone with a proper scoring rule
$\mathcal{L}_{\mathrm{pred}}
=\frac{1}{N}\sum_i
\mathcal{S}(p_\theta(\cdot\mid\mathbf{I}_i),\mathbf{Y}_i)$, where
$\mathcal{S}$ is the negative log-likelihood for explicit densities or CRPS
for sample-based forecasts. At inference, the model generates
\begin{equation}
\mathbf{Y}^{(k)}\sim p_\theta(\mathbf{Y}\mid\mathbf{I}),
\qquad k=1,\ldots,K,
\label{eq:forecast_samples}
\end{equation}
which preserves joint spatiotemporal uncertainty across variables and lead
times.

\subsection{Decision Risk Adapter}

For each action, the sampled outcomes induce losses
$\ell_a^{(k)}=L(a,\mathbf{Y}^{(k)};\mathbf{C})$. Their Monte Carlo mean
\begin{equation}
\bar r_K(a)=\frac{1}{K}\sum_{k=1}^{K}\ell_a^{(k)}
\label{eq:mc_risk}
\end{equation}
is a direct estimate of expected risk. Finite ensembles, model misspecification,
and context shift can nevertheless bias this estimate. We therefore use a
residual risk adapter
\begin{equation}
\widehat r_\psi(a\mid\mathbf{I},\mathbf{C})
=\bar r_K(a)+
\Delta_\psi\!\left(
\mathbf{h}_t,e(a),e(\mathbf{C}),
T(\{\ell_a^{(k)}\}_{k=1}^{K})
\right),
\label{eq:risk_adapter}
\end{equation}
where $e(\cdot)$ denotes an embedding and $T(\cdot)$ contains the empirical
mean, variance, and tail quantiles of sampled losses. The residual form retains
the decision-theoretic Monte Carlo estimate while allowing data-driven
correction.

Given observed outcomes, the adapter uses every feasible action as
supervision:
\begin{equation}
\mathcal{L}_{\mathrm{risk}}
=\frac{1}{N|\mathcal{A}|}
\sum_{i=1}^{N}\sum_{a\in\mathcal{A}}
\operatorname{Huber}\!\left(
\widehat r_\psi(a\mid\mathbf{I}_i,\mathbf{C}_i),
L(a,\mathbf{Y}_i;\mathbf{C}_i)
\right).
\label{eq:risk_loss}
\end{equation}
For risk-averse users, Eq.~\eqref{eq:mc_risk} is replaced by a weighted
combination of the sample mean and empirical CVaR. The resulting risk
preference is included in $\mathbf{C}$.

\subsection{Utility-Aware Calibration}

Forecast-level calibration does not guarantee reliable action losses. We use
a held-out calibration set
$\mathcal{D}_{\mathrm{cal}}=\{(\mathbf{I}_i,\mathbf{C}_i,\mathbf{Y}_i)\}_{i=1}^n$
and compute an action-specific loss-space score
\begin{equation}
s_i(a)=
L(a,\mathbf{Y}_i;\mathbf{C}_i)
-\widehat r_\psi(a\mid\mathbf{I}_i,\mathbf{C}_i).
\label{eq:cal_score}
\end{equation}
For each action, let $q_\alpha(a)$ be the
$\lceil(n+1)(1-\alpha/|\mathcal{A}|)\rceil/n$ empirical quantile of
$\{s_i(a)\}_{i=1}^{n}$. The calibrated upper risk is
\begin{equation}
\widetilde r_\alpha(a\mid\mathbf{I},\mathbf{C})
=\widehat r_\psi(a\mid\mathbf{I},\mathbf{C})+q_\alpha(a).
\label{eq:cal_risk}
\end{equation}
Under exchangeability of calibration and test examples, conformal calibration
gives
\begin{equation}
\Pr\!\left[
\forall a\in\mathcal{A}:\,
L(a,\mathbf{Y};\mathbf{C})
\leq \widetilde r_\alpha(a\mid\mathbf{I},\mathbf{C})
\right]\geq 1-\alpha.
\label{eq:simultaneous_guarantee}
\end{equation}

\subsection{Risk-Aware Decision Rule and Optimization}

The final action minimizes calibrated risk:
\begin{equation}
a^\star=\arg\min_{a\in\mathcal{A}}
\widetilde r_\alpha(a\mid\mathbf{I},\mathbf{C}).
\label{eq:calibrated_action}
\end{equation}
When an application specifies a risk budget $\tau(\mathbf{C})$, we first form
\begin{equation}
\mathcal{A}_{\mathrm{safe}}
=\{a\in\mathcal{A}:
\widetilde r_\alpha(a\mid\mathbf{I},\mathbf{C})
\leq\tau(\mathbf{C})\}.
\label{eq:safe_set}
\end{equation}
The system selects the lowest-cost action in $\mathcal{A}_{\mathrm{safe}}$; if
the set is empty, it abstains, requests inspection, or executes a predefined
conservative action. We optimize the predictive model and risk adapter with
\begin{equation}
\mathcal{L}
=\mathcal{L}_{\mathrm{pred}}
+\lambda_{\mathrm{risk}}\mathcal{L}_{\mathrm{risk}},
\label{eq:total_loss}
\end{equation}
and apply utility-aware calibration post hoc on data excluded from training.
Inference requires $K$ forecast samples and evaluates
$|\mathcal{A}|$ action losses, giving decision-layer complexity
$O(K|\mathcal{A}|)$.

\section{Experiments}

\subsection{Datasets and Decision Tasks}

We use ERA5 and WeatherBench 2 for multivariate Earth system forecasting and
ExtremeWeatherBench for evaluating high-impact events. We split all data
chronologically into training, validation, calibration, and test sets to
prevent future information leakage. The validation set supports model
selection, while the calibration set only estimates the risk correction and
never contributes to model training.

\noindent\ding{172}\ \textbf{Extreme precipitation warning.}
The actions are no warning, standard warning, and severe warning.

\noindent\ding{173}\ \textbf{Heatwave resource allocation.}
The actions are no response, local response, and full response.

\noindent\ding{174}\ \textbf{Renewable-energy dispatch.}
The actions specify different levels of reserve-energy allocation.

For an observed future state $\mathbf{Y}$, the decision loss is
\begin{equation}
L(a,\mathbf{Y};\mathbf{C})
=L_{\mathrm{event}}(a,\mathbf{Y};\mathbf{C})
+\lambda_{\mathrm{cost}}C(a),
\label{eq:task_loss}
\end{equation}
where $L_{\mathrm{event}}$ measures the cost of missed events, false alarms, or
energy shortfalls, and $C(a)$ denotes the operational cost of action $a$.

\subsection{Baselines}

We compare the following methods under the same forecasting backbone, action
set, and loss function:

\noindent\ding{51}\ \textbf{Point Forecast} selects an action from the
deterministic prediction.

\noindent\ding{51}\ \textbf{Sample Risk} minimizes the mean loss over forecast
samples.

\noindent\ding{51}\ \textbf{Forecast Conformal} calibrates uncertainty at the
forecast-variable level.

\noindent\ding{51}\ \textbf{Conformal Risk Control} controls a general
decision loss.

\noindent\ding{51}\ \textbf{Ours w/o Adapter} removes the decision risk
adapter.

\noindent\ding{51}\ \textbf{Ours w/o Calibration} removes utility-aware
calibration.

\noindent\ding{51}\ \textbf{Ours} uses both action-conditional risk adaptation
and utility-aware calibration.

\subsection{Evaluation Metrics}

We measure forecast quality with RMSE, CRPS, interval coverage, and interval
width. We evaluate decision quality with decision regret, missed-event rate,
false-alarm rate, selective risk, expected utility, and action-loss coverage.
Decision regret is
\begin{equation}
\operatorname{Regret}
=L(a^\star,\mathbf{Y};\mathbf{C})
-\min_{a\in\mathcal{A}}L(a,\mathbf{Y};\mathbf{C}),
\label{eq:decision_regret}
\end{equation}
where a lower value indicates a decision closer to the hindsight-optimal
action.

\subsection{Implementation Details}

The foundation model generates $K$ future-state samples for estimating action-conditional risk. Conformal correction is computed on an independent calibration split with target coverage $1-\alpha$. All methods use identical data splits and forecasting settings.

Models are trained for 20 epochs using AdamW, with learning rates of $2\times10^{-5}$ for pretrained components and $1\times10^{-4}$ for new modules. We use a batch size of 16, weight decay of 0.01, 5\% linear warm-up, gradient clipping at 1.0, and early stopping on validation CRPS with a patience of three epochs. Experiments run on four NVIDIA A100 80\,GB GPUs with seeds 13, 42, and 2026. We report means and 95\% confidence intervals and vary $K$, $\alpha$, and asymmetric decision costs in sensitivity experiments.

\section{Results}

We investigate four questions: \textbf{RQ1}, whether decision-oriented UQ reduces decision risk; \textbf{RQ2}, how it compares with forecast-level calibration; \textbf{RQ3}, the contributions of the risk adapter and calibration module; and \textbf{RQ4}, robustness to risk levels, ensemble sizes, and distribution shifts.

\subsection{Overall Performance}

Table~\ref{tab:main_results} reports average performance across three decision tasks. As all methods share the same forecasting backbone, RMSE and CRPS remain similar, while calibration and decision quality differ.

Compared with the strongest forecast-calibration baseline, our method reduces decision regret from 0.139 to 0.113 (18.7\%) and the missed-event rate from 14.2\% to 9.1\%, while increasing expected utility from 0.684 to 0.763 (11.6\%). Predictive and action-loss coverage reach 90.4\% and 90.2\%, respectively, near the 90\% target.

Forecast Conformal achieves only 84.9\% action-loss coverage despite reliable predictive coverage, while Conformal Risk Control reaches 89.6\% but remains weaker in regret and utility. For extreme-precipitation warning, our method reduces regret by 23.4\%, demonstrating its advantage under asymmetric high-impact losses.

\begin{table}[!t]
\centering
\caption{Overall performance averaged over three decision tasks. Values report
means and 95\% confidence intervals. Arrows indicate the preferred direction;
coverage metrics target 90\%. Best and second-best decision results are shown
in bold and underline, respectively.}
\label{tab:main_results}
\small
\setlength{\tabcolsep}{4pt}
\textbf{(a) Forecast quality}\\[2pt]
\begin{tabularx}{\textwidth}{@{}lYYY@{}}
\hline
\rowcolor{TableHeader}
Method & RMSE$\downarrow$ & CRPS$\downarrow$ & Coverage (\%) \\
\hline
Point Forecast
& $0.842{\pm}0.004$ & -- & -- \\
\rowcolor{TableStripe}
Sample Risk
& $0.842{\pm}0.004$ & $0.486{\pm}0.003$ & $84.7{\pm}0.8$ \\
Forecast Conformal
& $0.842{\pm}0.004$ & $0.486{\pm}0.003$ & $90.1{\pm}0.6$ \\
\rowcolor{TableStripe}
Conformal Risk Control
& $0.842{\pm}0.004$ & $0.486{\pm}0.003$ & $90.3{\pm}0.7$ \\
\rowcolor{AblationBg}
Ours w/o Adapter
& $0.842{\pm}0.004$ & $0.486{\pm}0.003$ & $90.2{\pm}0.6$ \\
\rowcolor{AblationBg}
Ours w/o Calibration
& $0.842{\pm}0.004$ & $0.486{\pm}0.003$ & $85.8{\pm}0.8$ \\
\rowcolor{OursBg}
Ours
& $0.842{\pm}0.004$ & $0.486{\pm}0.003$ & $90.4{\pm}0.6$ \\
\hline
\end{tabularx}

\vspace{6pt}
\textbf{(b) Decision errors}\\[2pt]
\begin{tabularx}{\textwidth}{@{}lYYY@{}}
\hline
\rowcolor{TableHeader}
Method
& Regret$\downarrow$
& Missed (\%)$\downarrow$
& False alarm (\%)$\downarrow$ \\
\hline
Point Forecast
& $0.201{\pm}0.009$
& $19.6{\pm}1.1$
& $14.5{\pm}0.9$ \\
\rowcolor{TableStripe}
Sample Risk
& $0.162{\pm}0.008$
& $16.8{\pm}0.9$
& $12.8{\pm}0.7$ \\
Forecast Conformal
& $0.139{\pm}0.007$
& $14.2{\pm}0.8$
& $11.7{\pm}0.6$ \\
\rowcolor{TableStripe}
Conformal Risk Control
& $0.145{\pm}0.006$
& $15.1{\pm}0.7$
& $\mathbf{10.9{\pm}0.6}$ \\
\rowcolor{AblationBg}
Ours w/o Adapter
& $0.129{\pm}0.006$
& $11.7{\pm}0.7$
& $11.5{\pm}0.5$ \\
\rowcolor{AblationBg}
Ours w/o Calibration
& $\underline{0.121{\pm}0.005}$
& $\underline{10.2{\pm}0.6}$
& $11.6{\pm}0.5$ \\
\rowcolor{OursBg}
Ours
& $\mathbf{0.113{\pm}0.005}$
& $\mathbf{9.1{\pm}0.5}$
& $\underline{11.2{\pm}0.5}$ \\
\hline
\end{tabularx}

\vspace{6pt}
\textbf{(c) Decision utility and calibration}\\[2pt]
\begin{tabularx}{\textwidth}{@{}lYY@{}}
\hline
\rowcolor{TableHeader}
Method & Utility$\uparrow$ & Loss coverage (\%) \\
\hline
Point Forecast
& $0.612{\pm}0.012$ & -- \\
\rowcolor{TableStripe}
Sample Risk
& $0.661{\pm}0.010$ & $82.4{\pm}0.9$ \\
Forecast Conformal
& $0.684{\pm}0.009$ & $84.9{\pm}0.8$ \\
\rowcolor{TableStripe}
Conformal Risk Control
& $0.679{\pm}0.008$ & $89.6{\pm}0.7$ \\
\rowcolor{AblationBg}
Ours w/o Adapter
& $0.738{\pm}0.008$ & $\underline{89.8{\pm}0.6}$ \\
\rowcolor{AblationBg}
Ours w/o Calibration
& $\underline{0.750{\pm}0.007}$ & $85.2{\pm}0.8$ \\
\rowcolor{OursBg}
Ours
& $\mathbf{0.763{\pm}0.007}$ & $\mathbf{90.2{\pm}0.6}$ \\
\hline
\end{tabularx}
\end{table}
\FloatBarrier

\subsection{Decision Calibration and Selective Risk}

\begin{figure}[t]
\begin{minipage}[t]{0.49\textwidth}
\centering
\includegraphics[width=\linewidth]{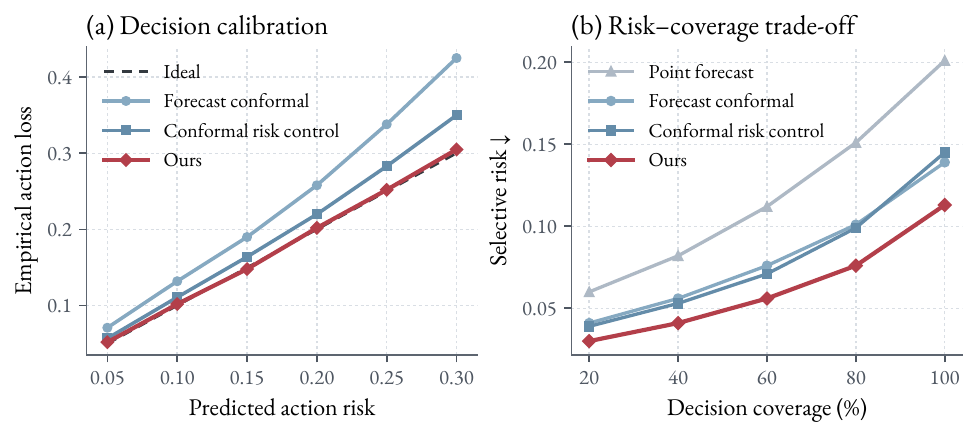}
\caption{Decision reliability analysis. (a) Predicted action risk versus
empirical loss. (b) Selective risk versus decision coverage.}
\label{fig:decision_calibration}
\end{minipage}
\hfill
\begin{minipage}[t]{0.49\textwidth}
\centering
\includegraphics[width=\linewidth]{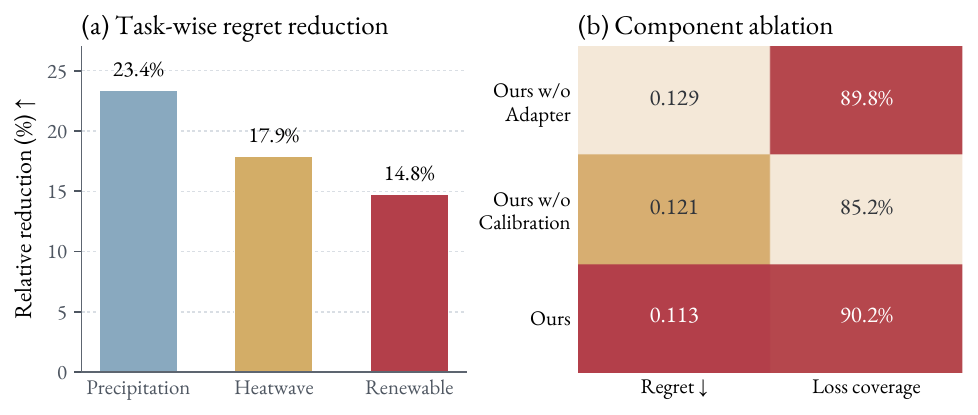}
\caption{Task and component analysis. (a) Task-wise regret reduction.
(b) Regret and action-loss coverage for model variants.}
\label{fig:decision_ablation}
\end{minipage}
\end{figure}

Figure~\ref{fig:decision_calibration}(a) shows that forecast-level calibration
increasingly underestimates empirical action loss as predicted risk grows.
At a predicted risk of 0.30, its empirical loss reaches 0.425, compared with
0.350 for Conformal Risk Control and 0.305 for our method. Accordingly, our
method reduces decision calibration error from 0.083 to 0.047 and remains close
to the ideal diagonal across the evaluated range.

Figure~\ref{fig:decision_calibration}(b) reports selective risk under different
decision coverage levels. Our method achieves the lowest risk throughout the
curve. At full coverage, it obtains a risk of 0.113, compared with 0.139 for
Forecast Conformal and 0.145 for Conformal Risk Control. The advantage remains
at lower coverage, indicating that utility-aware calibration improves both
automatic decisions and selective deployment.

\subsection{Ablation and Task-wise Analysis}

Figure~\ref{fig:decision_ablation}(a) shows consistent improvements across all
three tasks. Decision regret decreases by 23.4\% for precipitation warning,
17.9\% for heatwave response, and 14.8\% for renewable-energy dispatch. The
larger gain on precipitation warning reflects the value of action-conditional
risk modeling when missed events carry strongly asymmetric costs.

Figure~\ref{fig:decision_ablation}(b) separates the contributions of the two
core modules. Removing the risk adapter increases regret from 0.113 to 0.129,
showing that direct Monte Carlo risk does not fully correct finite-ensemble and
context-dependent errors. Removing utility-aware calibration increases regret
to 0.121 and reduces action-loss coverage from 90.2\% to 85.2\%. Thus, the
risk adapter primarily improves risk estimation, while calibration provides
reliable control of realized decision loss.

\subsection{Sensitivity and Robustness}

\begin{figure}[t]
\centering
\includegraphics[width=\textwidth]{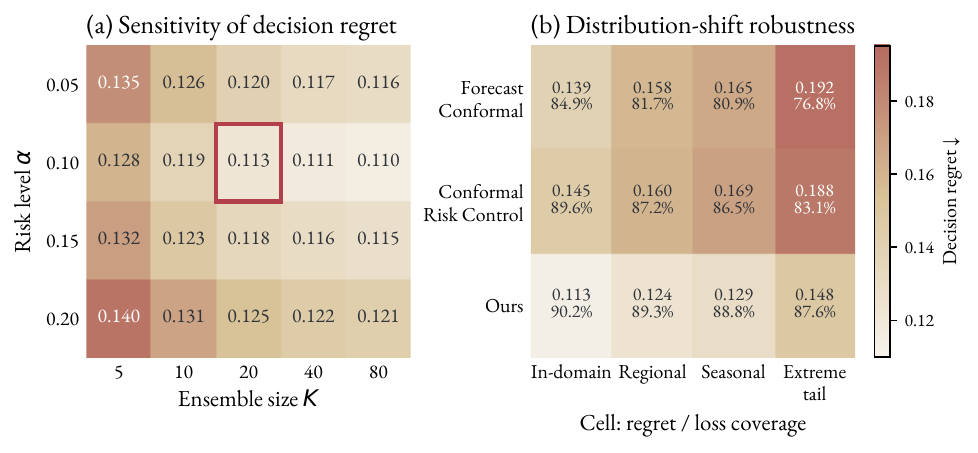}
\caption{Sensitivity and robustness analysis. (a) Decision regret for different
ensemble sizes $K$ and risk levels $\alpha$; the outlined cell is the default
setting. (b) Decision regret and action-loss coverage under regional, seasonal,
and extreme-tail shifts.}
\label{fig:decision_robustness}
\end{figure}

Figure~\ref{fig:decision_robustness}(a) shows that increasing the ensemble size
from $K=5$ to $K=20$ reduces regret from 0.128 to 0.113 at
$\alpha=0.10$. Beyond $K=20$, the improvement becomes marginal, indicating
that moderate ensembles capture most decision-relevant uncertainty. Across
risk levels, $\alpha=0.10$ provides the best balance: smaller values produce
overly conservative actions, whereas larger values permit more high-loss
decisions. We therefore use $K=20$ and $\alpha=0.10$ by default.

Figure~\ref{fig:decision_robustness}(b) evaluates robustness beyond the
in-domain test distribution. Under the extreme-tail shift, our method obtains
0.148 regret, compared with 0.192 for Forecast Conformal and 0.188 for
Conformal Risk Control. It also retains 87.6\% action-loss coverage, exceeding
the corresponding baselines by 10.8 and 4.5 percentage points. Regional and
seasonal shifts show the same trend. Although distribution shift weakens the
nominal conformal guarantee, utility-aware calibration degrades more
gracefully than forecast-level alternatives.

\subsection{Case Study: Marine Heatwave Advisory}

\begin{figure}[t]
\centering
\includegraphics[width=\textwidth]{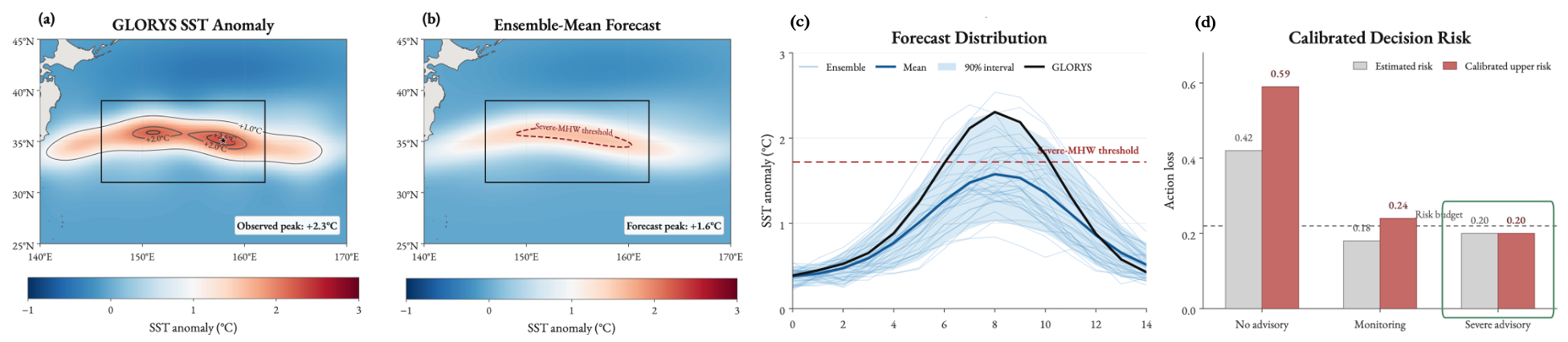}
\caption{Marine heatwave advisory in the Kuroshio Extension. (a) GLORYS SST
anomaly used for verification. (b) Ensemble-mean forecast. (c) Regional
forecast distribution and severe-MHW threshold. (d) Estimated and calibrated
risks for the three operational actions.}
\label{fig:mhw_case}
\end{figure}

Figure~\ref{fig:mhw_case} examines a held-out marine heatwave in the Kuroshio
Extension. The GLORYS verification shows an elongated warm anomaly with a peak
of $+2.3\,^{\circ}$C, whereas the ensemble-mean forecast reaches only
$+1.6\,^{\circ}$C and underestimates both its intensity and spatial extent.
Nevertheless, the ensemble distribution retains warm-tail trajectories that
cross the severe-MHW threshold, and its 90\% interval covers most of the
verified event evolution. This contrast illustrates why a deterministic mean
alone can conceal decision-relevant tail risk.

We consider three actions: no advisory, enhanced monitoring, and a severe-MHW
advisory. Their estimated losses are $\{0.42,0.18,0.20\}$, which would favor
monitoring before calibration. Utility-aware calibration produces upper risks
of $\{0.59,0.24,0.20\}$ by penalizing the asymmetric consequence of a missed
event. Under the risk budget, only the severe advisory remains reliably safe,
so the calibrated rule escalates the response. The case demonstrates how
GLORYS verification, ensemble tail information, and action-specific
calibration jointly convert an underestimated SST forecast into a
risk-appropriate operational decision.

\section{Conclusion}

This work presents a decision-oriented uncertainty quantification framework
for Earth system spatiotemporal foundation models. Rather than treating
forecast calibration as the final objective, the framework propagates
predictive samples through action-dependent loss functions, learns
context-aware decision risk, and calibrates that risk in utility space. The
resulting decision rule directly supports minimum-risk action selection,
risk-budget constraints, and abstention while retaining finite-sample
action-loss coverage under the stated exchangeability assumptions.

Across three representative Earth system decision tasks, the proposed method
reduces decision regret by 18.7\% relative to the strongest baseline, lowers
the missed-event rate from 14.2\% to 9.1\%, and improves expected utility by
11.6\%, while maintaining 90.2\% action-loss coverage. Ablation and
sensitivity analyses show that the risk adapter and utility-aware calibration
provide complementary gains, and the marine heatwave case illustrates how
ensemble tail information changes an operational response when the forecast
mean underestimates event severity. Future work will extend the framework to
continuous and sequential actions, calibration under nonstationary
distributions, and prospective evaluation with domain operators.

\bibliographystyle{splncs04}
\bibliography{ref}

@article{feng2026s,
  title={{S$^{2}$Q-VDiT$^{+}$}: Accurate Quantized Video Diffusion Transformer with Multi-Resolution Sampling and Structural Distillation},
  author={Feng, Weilun and Yang, Chuanguang and Qin, Haotong and others},
  journal={IEEE Transactions on Pattern Analysis and Machine Intelligence},
  year={2026},
  publisher={IEEE}
}

@article{feng2026mpq,
  title={Mpq-dmv2: Flexible residual mixed precision quantization for low-bit diffusion models with temporal distillation},
  author={Feng, Weilun and Yang, Chuanguang and Qin, Haotong and others},
  journal={IEEE Transactions on Pattern Analysis and Machine Intelligence},
  year={2026},
  publisher={IEEE}
}

@article{gou2021knowledge,
  title={Knowledge distillation: A survey},
  author={Gou, Jianping and Yu, Baosheng and Maybank, Stephen J and Tao, Dacheng},
  journal={International journal of computer vision},
  volume={129},
  number={6},
  pages={1789--1819},
  year={2021},
  publisher={Springer}
}

@article{gou2025multi,
  title={Multi-scale collaborative distillation graph neural networks for session-based recommendation},
  author={Gou, Jianping and Cheng, Youhui and Ma, Benteng and Du, Lan and Luo, Xin and Yi, Zhang},
  journal={IEEE Transactions on Services Computing},
  year={2025},
  publisher={IEEE}
}

@article{gou2022multilevel,
  title={Multilevel attention-based sample correlations for knowledge distillation},
  author={Gou, Jianping and Sun, Liyuan and Yu, Baosheng and Wan, Shaohua and Ou, Weihua and Yi, Zhang},
  journal={IEEE Transactions on Industrial Informatics},
  volume={19},
  number={5},
  pages={7099--7109},
  year={2022},
  publisher={IEEE}
}

@inproceedings{qi2026next,
  title={Next generation active learning: Mixture of llms in the loop},
  author={Qi, Yuanyuan and Yang, Xiaohao and Lu, Jueqing and Guo, Guoxiang and Enticott, Joanne and Liu, Gang and Du, Lan},
  booktitle={Proceedings of the AAAI Conference on Artificial Intelligence},
  volume={40},
  number={29},
  pages={24909--24917},
  year={2026}
}

@inproceedings{cheng2025cgmatch,
  title={Cgmatch: A different perspective of semi-supervised learning},
  author={Cheng, Bo and Lu, Jueqing and Tian, Yuan and Zhao, Haifeng and Chang, Yi and Du, Lan},
  booktitle={2025 IEEE/CVF Conference on Computer Vision and Pattern Recognition (CVPR)},
  pages={15381--15391},
  year={2025},
  organization={IEEE}
}

@article{li2026mrmad,
  title={MRMAD: A Multi-Round Multi-Audio Benchmark for Evaluating Acoustic Degradation Perception in Large Audio-Language Models},
  author={Li, Yize and Yang, Ningyuan and Yin, Sile and Thogarrati, Sindhuja and Chang, Sung-En and Singer, Andrew C and Lin, Xue and Huang, Chuan-Che and Zhang, Shuo},
  journal={arXiv preprint arXiv:2608.22236},
  year={2026}
}

@inproceedings{li2026diff,
  title={Diff-StyGS: 3D Gaussian Splatting Stylization via Tuning-Free Multi-view Sparse Diffusion},
  author={Li, Yize and Lu, Lei and Kong, Zhenglun and Wang, Yanzhi and Zhao, Pu and Lin, Xue},
  booktitle={International Conference on Pattern Recognition},
  pages={620--635},
  year={2026},
  organization={Springer}
}

@inproceedings{li2025pruning,
  title={Pruning then reweighting: Towards data-efficient training of diffusion models},
  author={Li, Yize and Zhang, Yihua and Liu, Sijia and Lin, Xue},
  booktitle={ICASSP},
  pages={1--5},
  year={2025},
  organization={IEEE}
}

@article{li2023less,
  title={Less is more: Data pruning for faster adversarial training},
  author={Li, Yize and Zhao, Pu and Lin, Xue and Kailkhura, Bhavya and Goldhahn, Ryan},
  journal={arXiv preprint arXiv:2302.12366},
  year={2023}
}

@article{li2024neural,
  title={Neural architecture search for adversarial robustness via learnable pruning},
  author={Li, Yize and Zhao, Pu and Ding, Ruyi and Zhou, Tong and Fei, Yunsi and Xu, Xiaolin and Lin, Xue},
  journal={Frontiers in High Performance Computing},
  volume={2},
  pages={1301384},
  year={2024},
  publisher={Frontiers Media SA}
}

@inproceedings{li2025frequency,
  title={Frequency-aligned knowledge distillation for lightweight spatiotemporal forecasting},
  author={Li, Yuqi and Yang, Chuanguang and Zeng, Hansheng and Dong, Zeyu and An, Zhulin and Xu, Yongjun and Tian, Yingli and Wu, Hao},
  booktitle={Proceedings of the IEEE/CVF International Conference on Computer Vision},
  pages={7262--7272},
  year={2025}
}

@ARTICLE{11541222,
  author={Li, Yuqi and Dong, Junhao and Liu, Jiao and Koniusz, Piotr and Zeng, Hansheng and Yang, Chuanguang and Liu, Junming and Tian, Yingli and Huang, Tingwen and Wu, Hao},
  journal={IEEE Transactions on Evolutionary Computation}, 
  title={Evolving Multimodal Models for Physical Dynamics: A Multi-objective Neuroevolution Approach}, 
  year={2026}
}

@article{li2026comprehensive,
  title={A Comprehensive Survey of Interaction Techniques in 3D Scene Generation},
  author={Li, Yuqi and Meng, Siwei and Yang, Chuanguang and Feng, Weilun and Liu, Junming and An, Zhulin and Wang, Yikai and Tian, Yingli},
  journal={IJCAI},
  year={2026}
}

@article{li2025ddtime,
  title={DDTime: Dataset Distillation with Spectral Alignment and Information Bottleneck for Time-Series Forecasting},
  author={Li, Yuqi and Ding, Kuiye and Yang, Chuanguang and Wang, Hao and Wang, Haoxuan and Duan, Huiran and Liu, Junming and Tian, Yingli},
  journal={arXiv preprint arXiv:2511.16715},
  year={2025}
}

@article{li2026rethinking,
  title={Rethinking Layer-Wise Information Allocation for Vision Foundation Model Adaptation},
  author={Li, Yuqi and Xiao, Xi and Zhang, Yunbei and Zhao, Lin and Li, Yu and Zhao, Aiden and Wang, Tianyang and Xu, Hao and Tian, Yingli},
  journal={arXiv preprint arXiv:2607.21973},
  year={2026}
}

@article{wu2026roboalign,
  title={RoboAlign-R1: Distilled Multimodal Reward Alignment for Robot Video World Models},
  author={Wu, Hao and Li, Yuqi and Gao, Yuan and Xu, Fan and Zhang, Fan and Wang, Kun and Zhao, Penghao and Wang, Qiufeng and Zhao, Yizhou and Wang, Weiyan and others},
  journal={arXiv preprint arXiv:2605.03821},
  year={2026}
}

@INPROCEEDINGS{11460474,
  author={Li, Yuqi and Ding, Kuiye and Yang, Chuanguang and Chen, Szu-Yu and Tian, Yingli},
  booktitle={ICASSP}, 
  title={Distilling Time Series Foundation Models for Efficient Forecasting}, 
  year={2026}
}

@inproceedings{xie2026symmetry,
  title={Symmetry-Aware Causal Inference for Robust Neural PDE Solvers},
  author={Xie, Yuanming and Xiang, Yanzhuo and You, Haochen and Liu, Nuoya and Liu, Fangzhou and Zhao, Bo and Kang, Zhaolu and Li, Yue and Li, Yuqi},
  booktitle={Proceedings of the 2026 International Conference on Multimedia Retrieval},
  pages={748--757},
  year={2026}
}

@article{Li2025Efficient,
  author  = {Li, Yuqi and Zeng, Hansheng and Zhang, Fuyan and Yang, Chuanguang and Li, Yanli and Ding, Weiping},
  title   = {{Efficient Medical Image Segmentation via Reinforcement Learning-Driven K-Space Sampling}},
  journal = {IEEE Transactions on Emerging Topics in Computational Intelligence},
  year    = {2025},
  publisher = {IEEE}
}

@article{zhao2026mis,
  title={MIS-HCC: Hierarchical Channel Clustering for Efficient Medical Image Segmentation},
  author={Zhao, Bo and Yu, Haoran and Liu, Lifei and Chu, Zongcheng and Liu, Yining and Liu, Chang and Chen, Szu-Yu and Xie, Zequn},
  journal={arXiv preprint arXiv:2607.17329},
  year={2026}
}

@article{bi2023pangu,
  author  = {Bi, K. and Xie, L. and Zhang, H. and Chen, X. and Gu, X. and Tian, Q.},
  title   = {Accurate Medium-Range Global Weather Forecasting with {3D} Neural Networks},
  journal = {Nature},
  volume  = {619},
  pages   = {533--538},
  year    = {2023}
}

@article{lam2023graphcast,
  author  = {Lam, R. and Sanchez-Gonzalez, A. and Willson, M. and others},
  title   = {Learning Skillful Medium-Range Global Weather Forecasting},
  journal = {Science},
  volume  = {382},
  number  = {6677},
  pages   = {1416--1421},
  year    = {2023}
}

@article{price2025gencast,
  author  = {Price, I. and Sanchez-Gonzalez, A. and Alet, F. and others},
  title   = {Probabilistic Weather Forecasting with Machine Learning},
  journal = {Nature},
  volume  = {637},
  pages   = {84--90},
  year    = {2025}
}

@article{bodnar2025aurora,
  author  = {Bodnar, C. and Bruinsma, W. P. and Lucic, A. and others},
  title   = {A Foundation Model for the {Earth} System},
  journal = {Nature},
  volume  = {641},
  pages   = {1180--1187},
  year    = {2025}
}

@article{ozdemir2026esfm,
  author  = {Ozdemir, F. and Cheng, Y. and Mohebi, S. and others},
  title   = {{Earth System Foundation Model (ESFM)}: A Unified Framework for Heterogeneous Data Integration and Forecasting},
  journal = {arXiv preprint arXiv:2605.00850},
  year    = {2026}
}

@article{yadav2026stochastic,
  author  = {Yadav, A. and Adebiyi, T. A. and Zhang, R.},
  title   = {Calibrating Scientific Foundation Models with Inference-Time Stochastic Attention},
  journal = {arXiv preprint arXiv:2604.19530},
  year    = {2026},
}

@inproceedings{minoza2026ntkuq,
  author    = {Mi{\~n}oza, J. M. A. and Laylo, R. G. and Iba{\~n}ez, S. C.},
  title     = {Scalable Uncertainty Quantification for Extreme Weather Forecasting via Empirical Neural Tangent Kernels},
  booktitle = {Proceedings of the 32nd ACM SIGKDD Conference on Knowledge Discovery and Data Mining},
  year      = {2026}
}

@article{asch2026conformal,
  author  = {Asch, A. and Rossellini, R. and Hassanzadeh, P. and Willett, R.},
  title   = {Rigorous Uncertainty Quantification of Probabilistic {AI} Weather Forecasts with Conformal Prediction},
  journal = {arXiv preprint arXiv:2606.19642},
  year    = {2026}
}

@article{schneider2026decision,
  author  = {Schneider, A. and Rochussen, T. and Stiller, J. and Fortuin, V.},
  title   = {Decision-Aligned Evaluation of Uncertainty Quantification},
  journal = {arXiv preprint arXiv:2606.26990},
  year    = {2026}
}

@article{angelopoulos2023conformal,
  author  = {Angelopoulos, A. N. and Bates, S.},
  title   = {Conformal Prediction: A Gentle Introduction},
  journal = {Foundations and Trends in Machine Learning},
  volume  = {16},
  number  = {4},
  pages   = {494--591},
  year    = {2023}
}

@inproceedings{wu2024earthfarsser,
  author    = {Wu, H. and Liang, Y. and Xiong, W. and Zhou, Z. and Huang, W. and Wang, S. and Wang, K.},
  title     = {{Earthfarsser}: Versatile Spatio-Temporal Dynamical Systems Modeling in One Model},
  booktitle = {Proceedings of the AAAI Conference on Artificial Intelligence},
  volume    = {38},
  number    = {14},
  pages     = {15906--15914},
  year      = {2024}
}

@inproceedings{gao2025oneforecast,
  author    = {Gao, Y. and Wu, H. and Shu, R. and others},
  title     = {{OneForecast}: A Universal Framework for Global and Regional Weather Forecasting},
  booktitle = {Proceedings of the 42nd International Conference on Machine Learning},
  series    = {Proceedings of Machine Learning Research},
  volume    = {267},
  pages     = {18658--18697},
  year      = {2025}
}

@article{wu2025tritoncast,
  author  = {Wu, H. and Gao, Y. and Gou, R. and others},
  title   = {Advanced Long-Term {Earth} System Forecasting},
  journal = {arXiv preprint arXiv:2505.19432},
  year    = {2025}
}

@article{xiong2023aigoms,
  author  = {Xiong, W. and Xiang, Y. and Wu, H. and Zhou, S. and Sun, Y. and Ma, M. and Huang, X.},
  title   = {{AI-GOMS}: Large {AI}-Driven Global Ocean Modeling System},
  journal = {arXiv preprint arXiv:2308.03152},
  year    = {2023}
}

@inproceedings{gao2026neuralom,
  author    = {Gao, Y. and Wu, H. and Xu, F. and Xiang, Y. and Gou, R. and Shu, R. and others},
  title     = {{NeuralOM}: Neural Ocean Model for Subseasonal-to-Seasonal Simulation},
  booktitle = {Proceedings of the AAAI Conference on Artificial Intelligence},
  volume    = {40},
  number    = {17},
  pages     = {14756--14764},
  year      = {2026}
}

@inproceedings{wu2026pnp,
  author        = {Wu, H. and Xu, F. and Lu, Y. and Zhao, P. and Zhang, F. and Jia, H. and others},
  title         = {{PnP-Corrector}: A Universal Correction Framework for Coupled Spatiotemporal Forecasting},
  booktitle     = {Proceedings of the 43rd International Conference on Machine Learning},
  series        = {PMLR},
  volume        = {306},
  year          = {2026},
  eprint        = {2605.08935},
  archiveprefix = {arXiv}
}

@article{xu2026tyche,
  author  = {Xu, F. and Gao, Y. and Wang, K. and Su, R. and Ling, F. and Wu, H. and Ouyang, W.},
  title   = {{Tyche}: One Step Flow for Efficient Probabilistic Weather Forecasting},
  journal = {arXiv preprint arXiv:2605.06916},
  year    = {2026}
}

@article{wu2025planning,
  author  = {Wu, H. and Gao, Y. and Shi, X. and Li, S. and Xu, F. and Zhang, F. and others},
  title   = {Spatiotemporal Forecasting as Planning: A Model-Based Reinforcement Learning Approach with Generative World Models},
  journal = {arXiv preprint arXiv:2510.04020},
  year    = {2025}
}

@article{shu2025mhw,
  author  = {Shu, R. and Wu, H. and Gao, Y. and Xu, F. and Gou, R. and Xiong, W. and Huang, X.},
  title   = {Advanced Forecasts of Global Extreme Marine Heatwaves through a Physics-Guided Data-Driven Approach},
  journal = {Environmental Research Letters},
  volume  = {20},
  number  = {4},
  pages   = {044030},
  year    = {2025}
}

@article{wang2026precip,
  author  = {Wang, Y. and Chen, H. and Wu, H. and Liu, J. and Yuan, H. and Cao, S. and Wang, T. and Zhuang, B.},
  title   = {Physics-Constrained Network for Enhanced Extended-Range Precipitation Forecasting in {East Asia}},
  journal = {Geophysical Research Letters},
  volume  = {53},
  number  = {6},
  pages   = {e2025GL120379},
  year    = {2026}
}

@inproceedings{wilder2019dfl,
  author    = {Wilder, B. and Dilkina, B. and Tambe, M.},
  title     = {Melding the Data-Decisions Pipeline: Decision-Focused Learning for Combinatorial Optimization},
  booktitle = {Proceedings of the AAAI Conference on Artificial Intelligence},
  volume    = {33},
  number    = {1},
  pages     = {1658--1665},
  year      = {2019}
}

@inproceedings{angelopoulos2024crc,
  author    = {Angelopoulos, A. N. and Bates, S. and Fisch, A. and Lei, L. and Schuster, T.},
  title     = {Conformal Risk Control},
  booktitle = {International Conference on Learning Representations},
  year      = {2024}
}
\end{document}